\documentclass[fleqn,10pt]{wlscirep}
\usepackage[utf8]{inputenc}
\usepackage[T1]{fontenc}

\usepackage{makecell}

\usepackage[acronym]{glossaries}
\makeglossaries
\newacronym{cnn}{CNN}{convolutional neural network}
\newacronym{ml}{ML}{machine learning}
\newacronym{ssl}{SSL}{self-supervised learning}
\newacronym{hts}{HTS}{high-throughput screening}
\newacronym{hpf}{hpf}{hours post fertilization}
\newacronym{imagenet}{IN1k}{ImageNet-1k}
\newacronym{mlp}{MLP}{multilayer perceptron}
\newacronym{umap}{UMAP}{uniform manifold approximation and projection}
\newacronym{vit}{ViT}{vision transformer}
\newacronym{knn}{k-NN}{k-nearest neighbor}

\title{Self-supervised learning for zebrafish embryo phenotype evaluation: Systematic comparison of algorithms, label-efficient classification, and transfer learning}

\author[1,*]{Thomas Lautenschlager}
\author[2,3]{Ga\"elle Hayot}
\author[2]{Quentin Brassart}
\author[1]{Angelo Jovin Yamachui Sitcheu}
\author[1]{Linus Schilpp}
\author[4,5]{Tijana Geroski}
\author[6]{Tetyana Goldanova}
\author[1]{Stephan Allgeier}
\author[6]{Thibaut Soubri\'e}
\author[4,7]{Milica Nikolic}
\author[2]{Thomas Dickmeis}
\author[1,+]{Ralf Mikut}
\author[1,8,+]{Nils Friederich}
\affil[1]{Karlsruhe Institute of Technology, Institute for Automation and Applied Informatics, Eggenstein-Leopoldshafen, 76344, Germany}
\affil[2]{Karlsruhe Institute of Technology, Institute for Biological and Chemical Systems - Biological Information Processing, Eggenstein-Leopoldshafen, 76344, Germany}
\affil[3]{Institut de Génétique et de Biologie Moléculaire et Cellulaire, Illkirch CEDEX, 67400, France}
\affil[4]{Steinbeis Advanced Risk Technologies Institute Kragujevac, Kragujevac, 34000, Serbia}
\affil[5]{University of Kragujevac, Faculty of Engineering, Kragujevac, 34000, Serbia}
\affil[6]{Preste, 242 boulevard Voltaire, Paris, 75011, France}
\affil[7]{University of Kragujevac, Faculty of Mechanical and Civil Engineering in Kraljevo, Kraljevo, 36000, Serbia}
\affil[8]{Karlsruhe Institute of Technology, Institute for Biological and Chemical Systems - Functional Molecular Systems, Eggenstein-Leopoldshafen, 76344, Germany}

\affil[*]{Corresponding author: thomas.lautenschlager@kit.edu}

\affil[+]{These authors jointly supervised this work: Ralf Mikut and Nils Friederich.}

\keywords{zebrafish, deep learning, self-supervised learning, transfer learning, toxicity, label efficiency}

\begin{abstract}
Zebrafish embryos are a prominent model organism for image-based phenotyping and especially suited for large-scale imaging studies, for which machine learning (ML) models can automate evaluation. However, training ML models usually requires large labeled datasets, and differences in imaging conditions often impede reuse across laboratories. Self-supervised learning (SSL) addresses both constraints by learning transferable representations from unlabeled images. Yet central aspects like the choice of algorithm remain underexplored for domain-specific applications. To close this gap, we benchmark popular SSL algorithms on the public EmbryoNet dataset, explore label-efficient learning, and test transfer to an unseen setting. Using representations from our strongest model, a ResNet50 network trained with SwAV, a simple linear classifier reaches 86.4\% accuracy and a non-linear classifier 87.1\%, rivaling the fully supervised EmbryoNet model (87.1\%). The SwAV model decisively outperforms supervised training from scratch when labels are scarce (70.2\% vs. 45.7\% accuracy with 5\% of labels). On a three-class problem (normal, dead, falsely cropped) with a small, previously unseen zebrafish toxicity dataset, recorded in a different setting, our model reaches 81.3\% accuracy without retraining, while EmbryoNet achieves 75.9\%. Overall, self-supervised models for zebrafish embryos seem to mostly facilitate label-efficiency and transfer to new settings rather than improving supervised results.
\end{abstract}
\begin{document}

\flushbottom
\maketitle
\thispagestyle{empty}

\section*{Introduction}
Image-based phenotyping enables close examination of how biological systems react to chemical or genetic perturbations, thereby deepening the understanding of how these systems work~\cite{caicedo2017data}. Research areas for which image-based phenotyping is central include developmental biology, drug discovery, and toxicology~\cite{capek2023embryonet, chandrasekaran2021image, seal2025cell}. Zebrafish (\textit{Danio rerio}) embryos are a prominent model organism in these contexts~\cite{capek2023embryonet, macrae2015zebrafish, cassar2020use, scholz2008zebrafish}, as they carry direct relevance for human biology, with 70\% of human genes having at least one obvious zebrafish orthologue~\cite{howe2013zebrafish}. The virtually transparent embryos develop externally, reaching phenotypically relevant stages within a day. This, combined with the high fecundity of zebrafish, makes them ideally suited for large-scale imaging. Using zebrafish embryos in drug discovery or toxicity tests is in line with the 3R principles that aim to replace, reduce, and refine animal experiments for more humane research~\cite{russell1959principles}.

Today, \gls{ml} can perform much of the evaluation of generated image data that would otherwise require trained experts~\cite{wang2024bringing}. For example, \v{C}apek et al. have shown that \Gls{ml} models can rival experienced researchers in zebrafish embryo phenotyping~\cite{capek2023embryonet}. Many such models have been proposed for the phenotyping of embryos~\cite{capek2023embryonet, sivaprasad2025automated} and already hatched embryos called eleutheroembryos that are not yet freely feeding~\cite{jeanray2015phenotype, shang2020zebrafish, dong2023deep, wang2025intelligent, tyagi2018fine}. All of these rely on supervised training, leading to practical limitations. First, supervised models need extensive amounts of labeled data for training. Especially in large-scale imaging, representative images can be acquired cheaply while expert labels are costly. Second, supervised models generally lose accuracy when imaging conditions change. Changes in microscope, illumination, or treatment protocols can degrade performance. This complicates sharing models between laboratories.

\Gls{ssl} addresses both limitations by learning transferable image representations without labels~\cite{uelwer2025survey}. A model first learns these general representations, compact numerical descriptions of an image, from raw, unlabeled images by solving an auxiliary task, for example, recognizing that two randomly cropped and distorted versions of the same embryo image belong together. The learned representations can then be used for training on downstream tasks such as classification or segmentation with far fewer labels. The practical outcome hinges on several choices, such as \gls{ssl} algorithm, pretraining strategy, and augmentations, i.e., random image modifications such as flips, rotations, or blur that are applied during training, which remain poorly understood for most biological domains. \Gls{ssl} algorithms vary in their auxiliary task for learning representations, making the choice of algorithm crucial. The pretraining strategy also plays a substantial role, namely whether to pretrain on a large natural-image corpus, e.g., \gls{imagenet}~\cite{deng2009imagenet}, or on domain images. Augmentations are critical as well, since the learning signal in most methods is tightly coupled to comparisons between differently augmented views of an image~\cite{uelwer2025survey}. Which specific augmentations are suited for a given domain is rarely studied. 

\Gls{ssl} algorithms are usually benchmarked on natural-image corpora such as \gls{imagenet}~\cite{deng2009imagenet}. Biomedical studies comparing \gls{ssl} algorithms have mostly focused on medical imaging, histopathology, or cellular microscopy rather than whole-organism phenotyping~\cite{huang2024systematic, kang2023benchmarking, kim2025self}. In zebrafish embryo images, discriminative phenotypic differences can be subtle, fine-grained, and spatially localized~\cite{capek2023embryonet}. A systematic comparison is therefore needed for applying \gls{ssl} to this imaging domain with confidence.

Three studies have examined representation learning without labels for zebrafish embryo images to our knowledge. Toulany et al. propose a model able to compare embryos against normal developmental trajectories~\cite{toulany2023uncovering}. Oktay et al. train a model on multiple datasets containing zebrafish embryo and eleutheroembryo images for use on several downstream tasks~\cite{oktay2026self}. A previous proof of concept of ours shows that SimCLR~\cite{chen2020simple}, a popular \gls{ssl} algorithm, can be applied for toxicity testing using zebrafish embryos~\cite{lautenschlager2025self}. None of these works deliver a systematic comparison of \gls{ssl} algorithms and architectures, explore label-efficient learning, or test transfer to a new out-of-domain setting.

In this work, we address these gaps by providing a comprehensive study of \gls{ssl} for zebrafish embryo phenotype classifications on the public EmbryoNet dataset~\cite{embryonetdata}. We discuss the results and derive actionable recommendations for applying \gls{ssl} across downstream zebrafish embryo tasks. Additionally, we release the most promising pretrained models for future research. 

\section*{Results} 

\begin{figure}[t!]
\centering
\includegraphics[width=\linewidth]{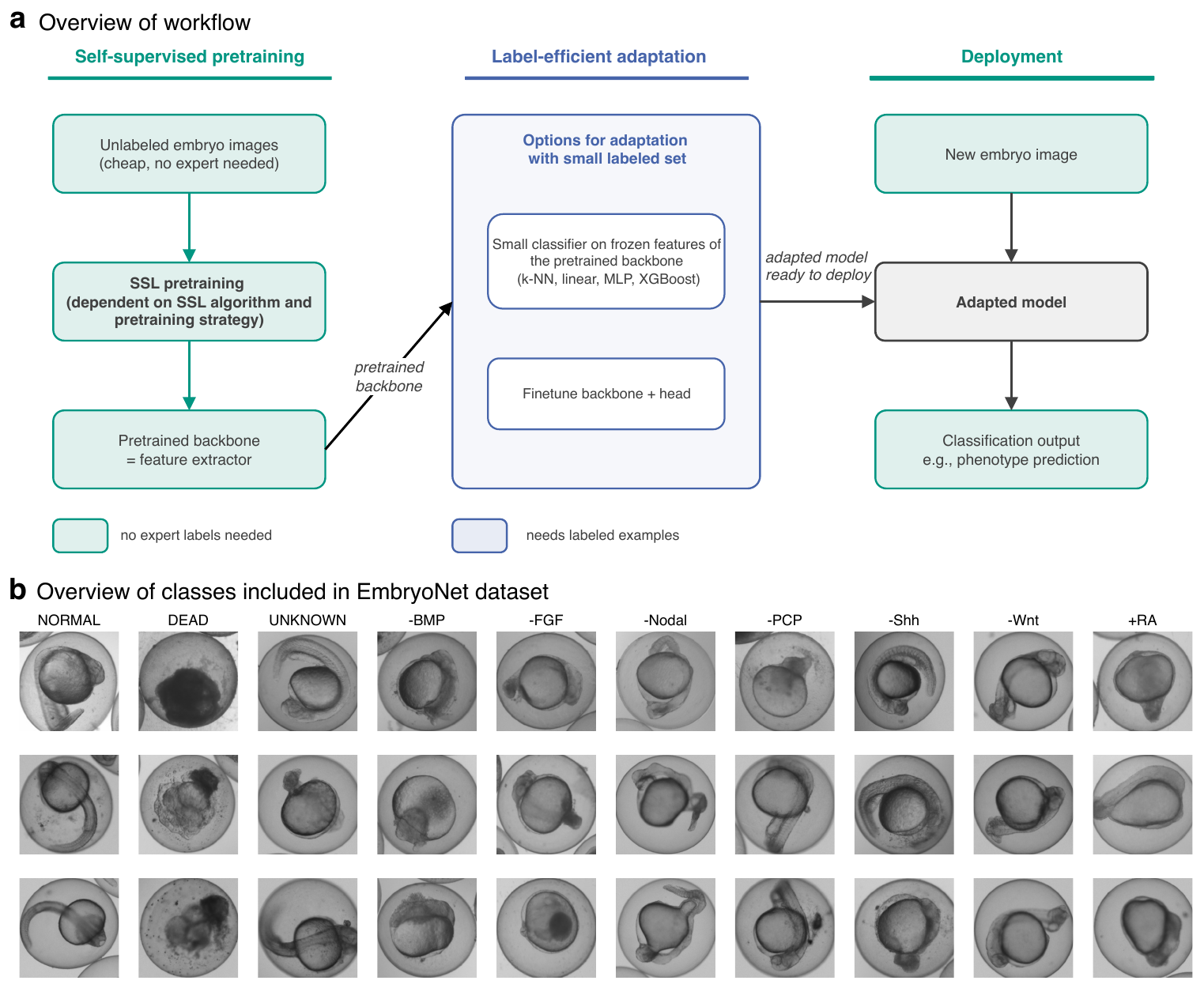}
\caption{\textbf{Overview of (a) the example workflow of a model in this study and (b) the EmbryoNet dataset.} (\textbf{a})~Self-supervised pretraining: A feature extractor is trained using SSL on unlabeled embryo images of the public EmbryoNet dataset and without expert labels. Label-efficient adaptation: The frozen features of the pretrained backbone are reused for adaptation to a specific task. Options are training small classifiers like k-NN, linear, MLP, or XGBoost on top of the frozen representations or full finetuning of the backbone. Deployment: The fully trained model is deployed for classification on unseen images, e.g., phenotype prediction. (\textbf{b})~Example images for each of the ten phenotypes in the EmbryoNet~\cite{embryonetdata} dataset were chosen at random from the test dataset. NORMAL = normally developing embryos, DEAD = dead/coagulated embryos, UNKNOWN = embryos that annotators could not clearly identify. The remaining labels indicate phenotypes resulting from dysfunctional major signaling pathways, with "-" indicating loss of function and "+" indicating gain of function of the pathway. The affected pathways are: BMP (Bone Morphogenetic Protein), FGF (Fibroblast Growth Factor), Nodal, PCP (Planar Cell Polarity), Shh (Sonic Hedgehog), Wnt (Wingless-related integration site), and RA (Retinoic Acid) signaling.}
\label{fig:methods}
\end{figure}

Our study follows a five-step study plan, in which the outcome of each step determines which models are carried forward to the next. First, we test different network architectures in a supervised setting. Second, we benchmark popular \gls{ssl} algorithms with the best networks from the first step. Third, we explore different pretraining strategies and how they affect label-efficient learning with the best combination of backbone and \gls{ssl} algorithm from the second step. Fourth, we analyze how individual augmentations impact the learned representations. Fifth, we investigate how well a pretrained model transfers to a new laboratory setting with a different microscope. Supplementary Figure~\ref{fig:testplan} gives an overview of the study plan. Figure~\ref{fig:methods}a depicts the workflow of an exemplary model, from label-free pretraining through label-efficient adaptation to transfer to a new setting. 

We use the EmbryoNet dataset published by \v{C}apek et al.~\cite{embryonetdata} for our study. The dataset contains ten zebrafish embryo phenotypes: normally developing embryos, dead embryos, embryos with an unclear phenotype, and seven phenotypes caused by chemically blocking or activating a major developmental signaling pathway. For all tests, we use the images from the predefined test dataset and restrict our evaluations to the endpoint at 26 \gls{hpf}. As a training dataset, we use images from 25 \gls{hpf} to 26 \gls{hpf} from the predefined training dataset, resulting in a dataset of 135,484 images. We do this to retain a reasonably sized dataset while still training on images of a comparable developmental stage. Figure~\ref{fig:methods}b shows example images of the dataset. Handling of the dataset is depicted in Supplementary Figure~\ref{fig:datasplit}.

\subsection*{Step 1: Supervised architecture comparison}
In the first step, we compare different network architectures trained fully supervised. All results are depicted in Figure~\ref{fig:scatter} and listed in Supplementary Table~\ref{tab:supervised}. The published EmbryoNet model reaches 87.1\% accuracy when applied to the images of the test dataset at 26 \gls{hpf}. For the original setup, which classifies embryos over their whole development and uses a transition logic to smooth classifications over the development span, an accuracy of 89\% is reported~\cite{capek2023embryonet}.

\begin{figure}[t!]
\centering
\includegraphics[width=\linewidth]{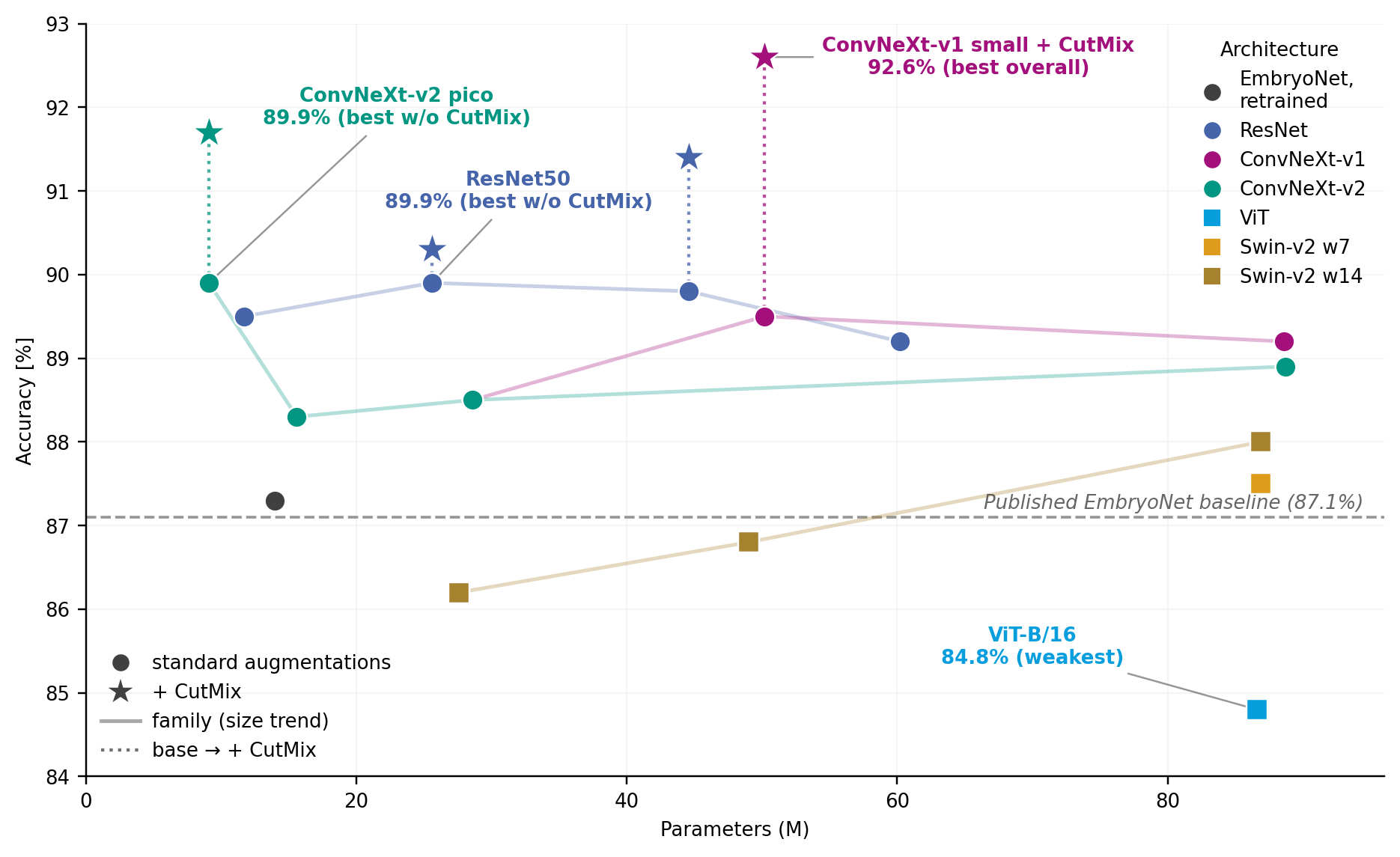}
\caption{\textbf{Accuracy of the fully supervised networks on the test dataset as a function of the number of model parameters (in millions - M).} Each point represents an architecture trained from scratch on the full training set. Solid lines connect the variants of one family in order of size. Dotted vertical lines link base models to their CutMix variant. The dashed line marks the published EmbryoNet baseline.}
\label{fig:scatter}
\end{figure}

Training convolutional and transformer architectures from scratch on the training dataset improves on this baseline. We test ResNets~\cite{he2016deep}, the ConvNeXt-v1s~\cite{liu2022convnet} and ConvNeXt-v2s~\cite{woo2023convnext}, vanilla \glspl{vit}~\cite{dosovitskiy2021image}, and Swin-v2 transformers~\cite{liu2022swin} of various sizes. ResNet50 and ConvNeXt-v2 pico reach the highest accuracy of 89.9\%. ConvNeXt-v2 pico achieves this with only 9.1 million parameters. The tested transformer architectures perform consistently worse. The best variant, Swin-v2 base w14, reaches 88.0\%. The vanilla \gls{vit}-B/16 trails the field at 84.8\%. We add CutMix~\cite{yun2019cutmix}, an augmentation technique that pastes an image patch from one image onto another and blends the labels proportionally, to the augmentations of the four best-performing architectures. It improves accuracy for all models with improvements ranging from 0.4 to 3.1 percentage points. ConvNeXt-v1 small yields the best overall result at 92.6\%. However, CutMix does not combine straightforwardly with most \gls{ssl} algorithms since the auxiliary task would be meaningfully altered by including patches from other images. We therefore carry ResNet50 and ConvNeXt-v2 pico without CutMix into the second step.

\begin{figure}[t!]
\centering
\includegraphics[width=\linewidth]{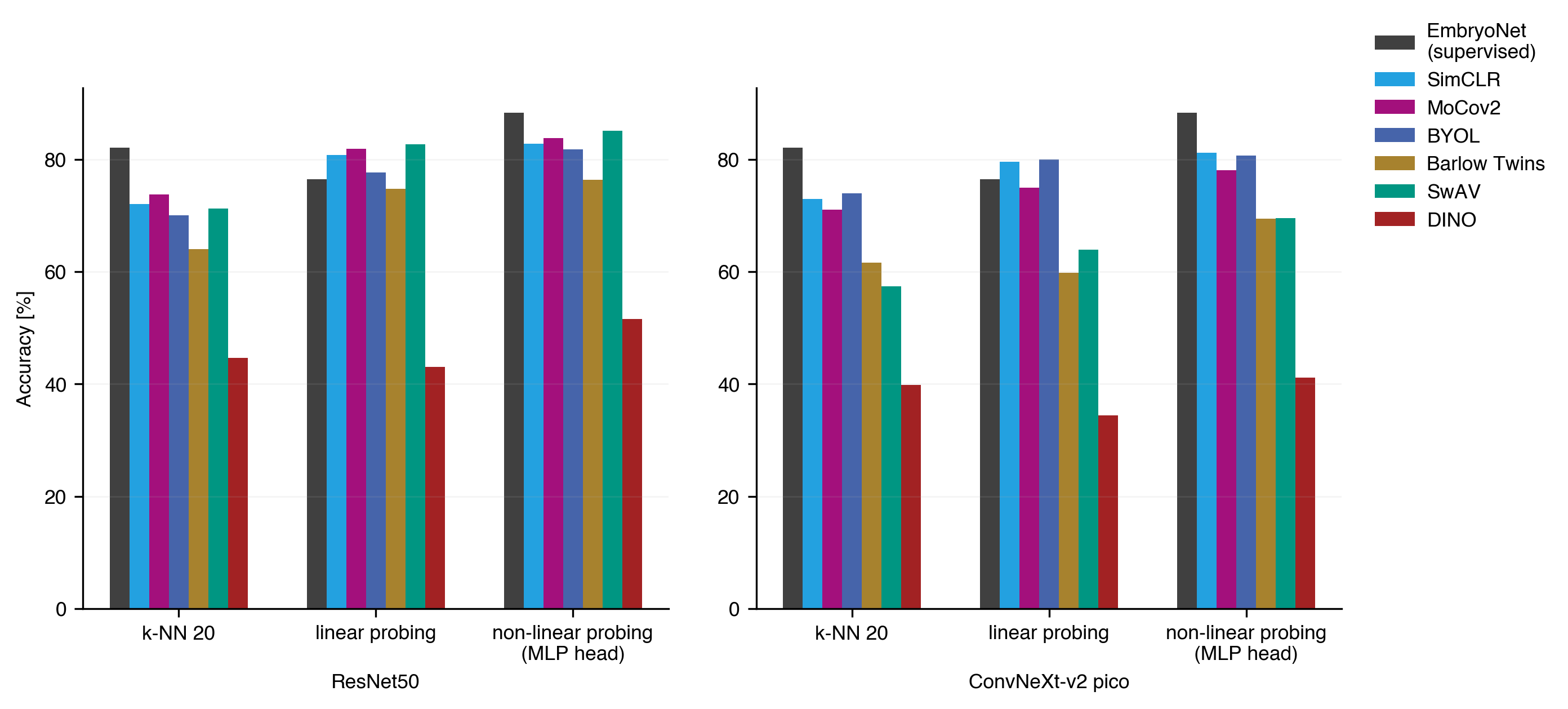}
    \caption{\textbf{Evaluation of the learned representations (200 epochs pretraining on the training dataset) of the self-supervised algorithms with ResNet50 and ConvNeXt-v2 pico backbones on the test dataset.} Accuracy is determined through k-NN classification with a k of 20 neighbors, linear probing, and non-linear probing with an MLP head on the frozen backbones. The representations of the published EmbryoNet model~\cite{embryonetdata} obtained by removing its classification head serve as a baseline. All results, including k-NN classifications with different numbers of neighbors, are given in Supplementary Table~\ref{tab:selfsupervised}.}
\label{fig:selfsupervised}
\end{figure}
 
\subsection*{Step 2: Self-supervised algorithm comparison}
In the second step, we compare six state-of-the-art \gls{ssl} algorithms for \glspl{cnn} using ResNet50 and ConvNeXt-v2 pico as backbones, i.e., the representation-extracting parts of the networks. The \gls{ssl} algorithms differ in the auxiliary task used for learning meaningful representations: SimCLR~\cite{chen2020simple} and MoCov2~\cite{chen2020improved} pull the representations of two differently augmented views of the same image together and push representations of other images away from them (contrastive learning). BYOL~\cite{grill2020bootstrap} and Barlow Twins~\cite{zbontar2021barlow} only pull representations of the same image together and prevent the trivial solution of mapping all representations to one single point in other ways. SwAV~\cite{caron2020unsupervised} requires representations of both views to be assigned to the same learned cluster, and DINO~\cite{caron2021emerging} lets a student network imitate a slowly updated copy of itself (self-distillation). More detailed descriptions of the algorithms are given in the Supplementary Methods. 

Each model is trained from scratch without expert labels for 200 epochs, i.e., 200 passes over the training set. We use \gls{knn} classification and linear probing, established evaluation methods in the \gls{ssl} field, to evaluate the learned representations. Linear probing assesses the linear separability of the learned representations. Further, we use a \gls{mlp} head for non-linear probing to determine non-linear separability of the representations. All of these classifiers are trained on top of the frozen backbones, i.e., the fixed representation-extracting networks that are not further trained, using the entire labeled training dataset.

The ResNet50 backbone outperforms ConvNeXt-v2 pico across almost all \gls{ssl} algorithms. Results of the \gls{ssl} comparison are visualized in Figure~\ref{fig:selfsupervised}, and all results are reported in Supplementary Table~\ref{tab:selfsupervised}. With ResNet50 backbones, SwAV performs best, achieving 82.7\% linear probing accuracy and 85.1\% with non-linear probing. SimCLR and MoCov2 follow closely with linear probing accuracies of 80.8\% and 81.9\%, respectively. BYOL and Barlow Twins reach 77.7\% and 74.8\% on linear probing. DINO performs considerably worse at 43.1\%. Using a ConvNeXt-v2 pico backbone only improves performance for BYOL, which reaches 80.0\% accuracy with linear probing, staying behind SimCLR, MoCov2, and SwAV with ResNet50 backbones. We further evaluate comparable representations from the published EmbryoNet model~\cite{embryonetdata} obtained by removing its classification head. The fully supervised model provides a strong baseline. It achieves 76.5\% with linear probing and reaches 88.3\% on non-linear probing. However, it was trained using all images with labels from the original EmbryoNet training dataset~\cite{embryonetdata}. Our SwAV model surpasses this reference on linear probing (82.7\% vs. 76.5\%) but not on non-linear probing (85.1\% vs. 88.3\%).

\begin{table}[b]
    \begin{tabular*}{\linewidth}{@{\extracolsep{\fill}}llccccccc}
     & \makecell[l]{training\\epochs} & \makecell[l]{k-NN\\1} &\makecell[l]{k-NN\\5} & \makecell[l]{k-NN\\10} & \makecell[l]{k-NN\\20} & \makecell[l]{k-NN\\200} & \makecell[l]{linear\\probing} & \makecell[l]{non-linear\\probing\\(MLP head)}\\
     \hline
     Supervised ImageNet & 100 & 54.3\% & 56.7\% & 56.1\% & 56.7\% & 60.3\% & 62.0\% & 71.1\%\\
     SwAV ImageNet & 800 & 54.3\% & 55.8\% & 56.2\% & 56.6\% & 58.6\% & 53.9\% & 65.2\%\\
     SwAV EmbryoNet & 800 & \textbf{75.5\%} & \textbf{76.3\%} & \textbf{75.9\%} & \textbf{76.4\%} & \textbf{78.9\%} & \textbf{86.4\%} & \textbf{87.1\%}\\
     SwAV ImageNet $\rightarrow$ EmbryoNet & 800 $\rightarrow$ 200 & 69.3\% & 70.7\% & 71.1\% & 72.0\% & 76.7\% & 83.0\% & 82.7\% \\
     \hline
    \end{tabular*}
    \caption{\textbf{Evaluation of different pretraining strategies using SwAV with a ResNet50 backbone on the test dataset.} Pretraining strategies use ImageNet-1k, EmbryoNet, or both sequentially. A ResNet50 pretrained fully supervised on ImageNet-1k serves as a baseline. The best results are marked in bold.}
    \label{tab:unsupervisedevaluation}
\end{table}
 
\subsection*{Step 3: Pretraining strategies and label-efficient classification}
In the third step, we compare three pretraining strategies using the best-performing model from the second step, SwAV with a ResNet50 backbone. We test SwAV pretraining using either only the natural image corpus \gls{imagenet} or only the domain dataset, EmbryoNet. In the third scenario, we sequentially pretrain the model first on \gls{imagenet} followed by EmbryoNet. Initial pretrainings are extended to 800 epochs. As a baseline, we use a fully supervised model trained with \gls{imagenet}. 

\begin{figure}[t!]
\centering
\includegraphics[width=\linewidth]{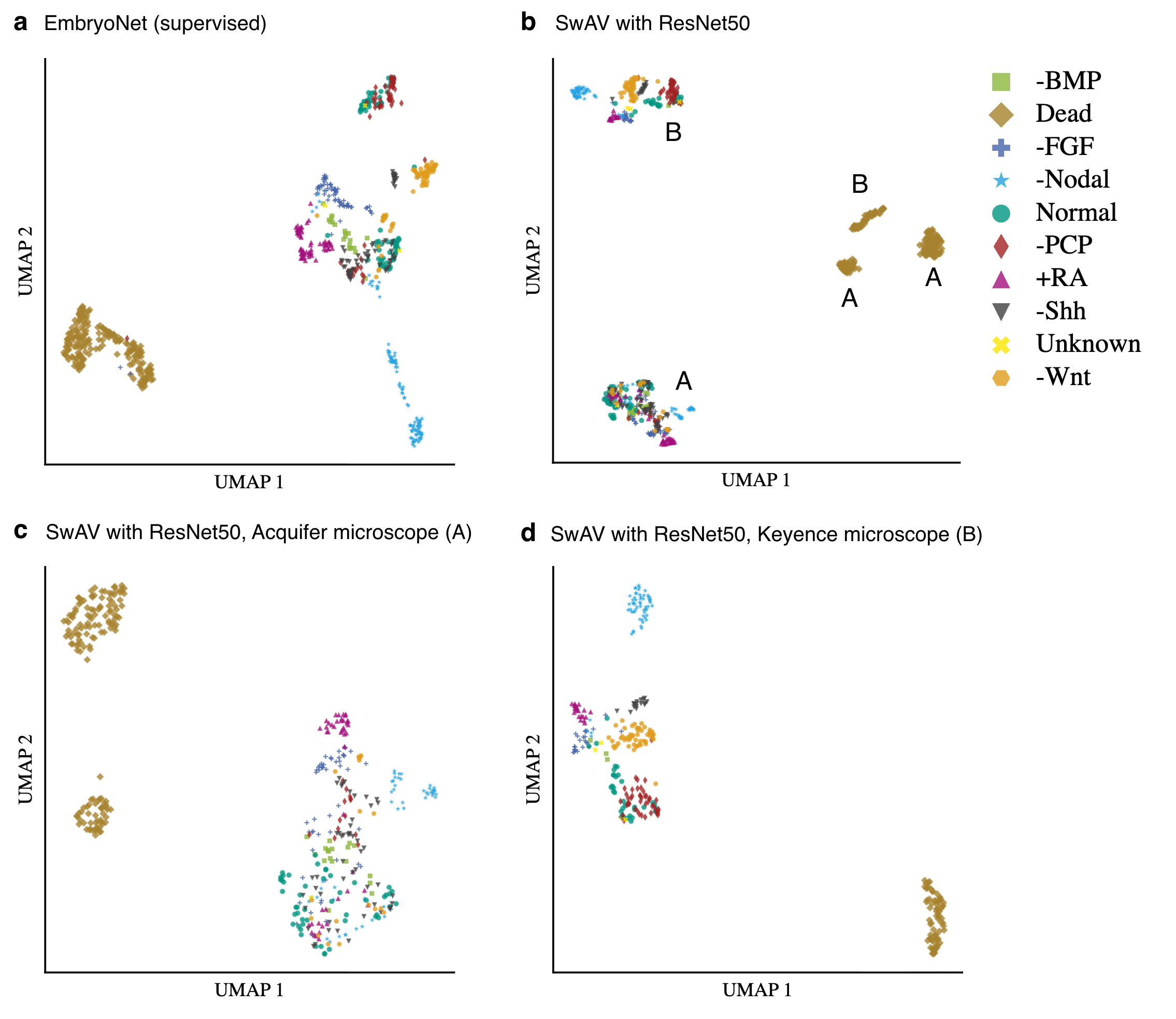}
\caption{\textbf{UMAP~\cite{mcinnes2018umap} projections of the learned representations from the test data of the published EmbryoNet model~\cite{embryonetdata} with the classification head removed (\textbf{a}) and of SwAV with a ResNet50 backbone (\textbf{b}, \textbf{c}, \textbf{d}).} SwAV separates based on the microscopes used for acquisition (Panel (\textbf{b}), A = Acquifer, B = Keyence). Separate UMAPs for Acquifer and Keyence are shown in Panel (\textbf{c}) and (\textbf{d}), respectively. SwAV was trained for 800 epochs on the EmbryoNet dataset.}
\label{fig:umaps}
\end{figure}

Results for the different pretraining strategies are given in Table~\ref{tab:unsupervisedevaluation}. Pretraining on EmbryoNet alone performs best across all evaluations. This strategy reaches 86.4\% via linear probing and 87.1\% via non-linear probing. Much of this improvement is driven by the extended pretraining. The linear probing accuracy rises from 82.7\% for 200 epochs of pretraining (cf. Figure~\ref{fig:selfsupervised} or Supplementary Table~\ref{tab:selfsupervised}) to 86.4\% for 800 epochs of pretraining. The sequential strategy ranks second with a linear probing accuracy of 83.0\%. Pretraining solely on \gls{imagenet} performs considerably worse, reaching only 53.9\%. The distinct clustering behavior of the representations is shown in Figure~\ref{fig:umaps}, visualized through dimension reduction via \gls{umap}~\cite{mcinnes2018umap}. The fully supervised EmbryoNet model~\cite{embryonetdata} with the classification head removed (Figure~\ref{fig:umaps}a) serves as a reference and largely separates the phenotype classes. The SwAV model (Figure~\ref{fig:umaps}b) organizes the representations primarily by the microscope used for acquisition. Within the data from a single microscope, the phenotype classes are reasonably separated (Figure~\ref{fig:umaps}c \& d), even without labels during training.

Further, we explore label-efficient training using the different SwAV pretraining strategies by finetuning, i.e., continuing to train the whole pretrained network on the labeled images. For this, we add a linear classification head to the backbone and train on increasing amounts of labeled data. An additional supervised baseline trained from scratch, with only the available labeled data, is added. We use 5\%, 10\%, 20\%, 50\%, and 100\% of the labeled training set and test different learning rates for finetuning. These fractions relate to roughly 168, 337, 674, 1684, and 3368 individual embryo development timelines with labels.

The results are visualized in Figure~\ref{fig:finetuning} and given in Supplementary Table~\ref{tab:finetuning}. Detailed results of the individual runs are given in Supplementary Table~\ref{tab:all_runs}. The EmbryoNet-pretrained model reaches 70.2\% using only 5\% of the labels, corresponding to labeled images of roughly 168 individual embryos. The sequentially trained model follows with an accuracy of 66.8\%. Pretraining the model only on \gls{imagenet} yields an accuracy of 58.8\%. The supervised \gls{imagenet} pretraining strategy achieves only 27.1\%, while training from scratch yields 45.7\%. Although the advantage of the \gls{ssl} models shrinks as more labels become available, the EmbryoNet-pretrained model remains the best model across all evaluations with limited labeled data. 

With 50\% of labeled data, the EmbryoNet-pretrained model reaches an accuracy of 88.0\%. This score closely rivals the 89.9\% of the fully supervised ResNet50 trained on all training labels from the first training step (cf. Figure~\ref{fig:scatter} or Supplementary Table~\ref{tab:supervised}) with only half the labels. Further, \gls{ssl}-pretrained models show much smaller standard deviations during finetuning across different learning rates.
 
\begin{figure}[t!]
\centering
\includegraphics[width=\linewidth]{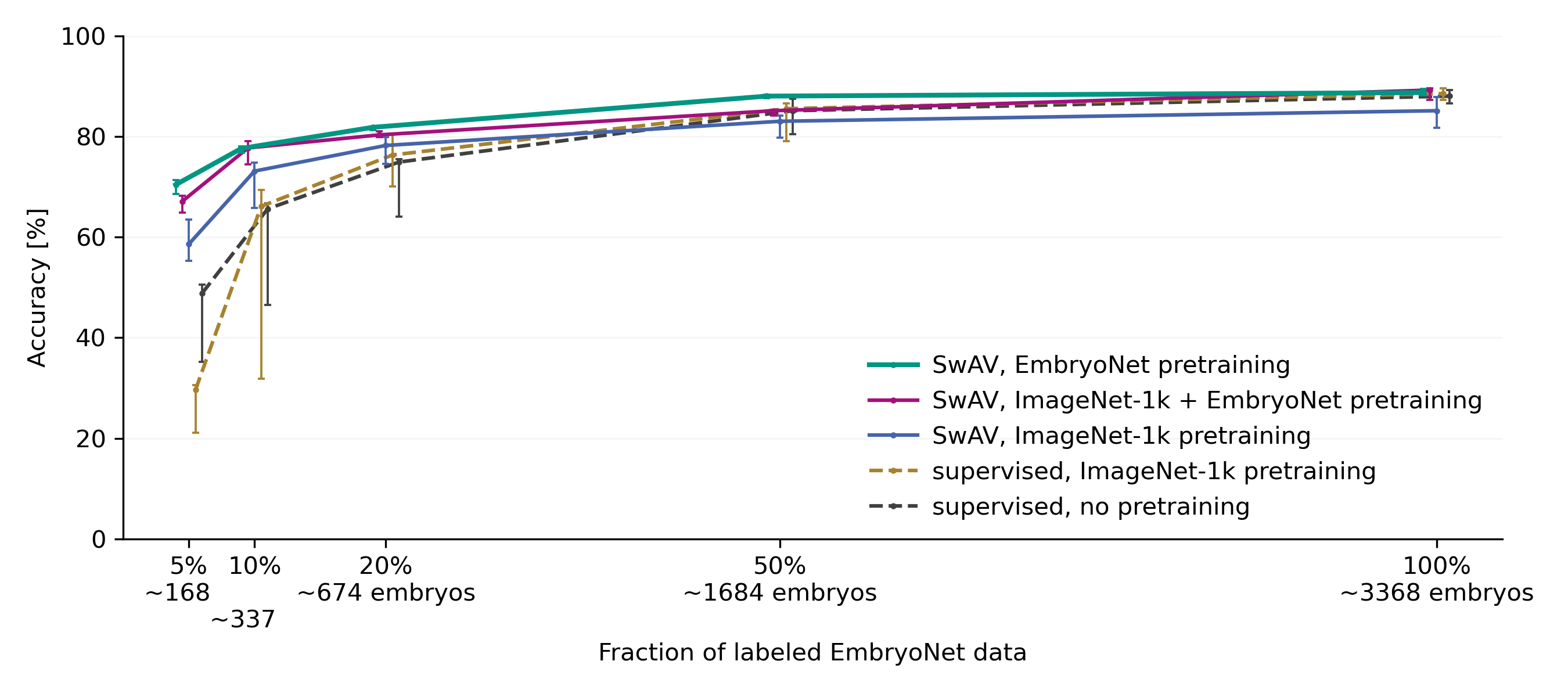}
\caption{\textbf{Mean accuracies (and standard deviations across five different learning rates) of the label-efficient training on the test dataset.} Backbones of the different self-supervised pretraining strategies are finetuned with labeled data fractions of 5\%, 10\%, 20\%, 50\%, and 100\% of the EmbryoNet dataset. Supervised backbones trained from scratch or with IN1k pretraining serve as baselines. Two of the five training runs with the supervised IN1k pretraining using 5\% labels did not converge. These runs were excluded from mean and standard deviation calculation. All results are given in Supplementary Table~\ref{tab:finetuning}. Results of individual training runs with the different learning rates are reported in Supplementary Table~\ref{tab:all_runs}.}
\label{fig:finetuning}
\end{figure}
 
\subsection*{Step 4: Augmentation study}
In the fourth step, we evaluate how the augmentations in our image augmentation pipeline affect the learned representations. We start with random cropping and add one augmentation at a time. All used augmentations with parameters are listed in Supplementary Table \ref{tab:augmentations_list}. We keep the SwAV and ResNet50 configuration from the second test step. 

The results are given in Table~\ref{tab:augmentations}. Adding augmentations generally improves performance. This is most prominent for the first few augmentations added, where performance improves over all metrics by large margins for most augmentations. However, substantial fluctuations in performance exist, and adding a single augmentation can also lead to deterioration for one or several of the evaluation methods. Generally, after a certain number of augmentations is applied, adding further augmentations does not improve results meaningfully. A certain number of augmentations appears to be necessary before the performance curve flattens. In addition, there appears to be no uniformly optimal augmentation combination, as the best-performing configurations vary based on the evaluation method used.
 
\begin{table}[t]
    \begin{tabular*}{\linewidth}{@{\extracolsep{\fill}}lcccc}
     Augmentations & k-NN 20 & k-NN 200 & linear probing & non-linear probing (MLP head) \\
         \hline
        only crops & 53.1\% & 56.7\% & 72.1\% & 73.0\% \\
        + horizontal flip & 56.2\% & 58.4\% & 71.7\% & 71.7\% \\
        + rotation & 63.3\% & 65.9\% & 76.6\% & 77.0\% \\
        + random brightness contrast changes & 70.0\% & 73.6\% & 79.5\% & 79.9\% \\
        + random gamma correction & 66.8\% & 69.9\% & 79.0\% & 81.3\% \\
        + CLAHE & 71.4\% & 73.8\% & 79.4\% & 79.1\% \\
        + sharpen & 72.5\% & 74.8\% & 84.1\% & 84.1\% \\
        + motion blur & 73.9\% & 78.4\% & \textbf{85.7\%} & 85.5\% \\
        + defocus blur & 72.7\% & 78.5\% & 85.0\% & 84.1\%\\
        + grid distortion & 73.9\% & 76.6\% & 84.4\% & 85.7\% \\
        + optical distortion & 72.9\% & 75.3\% & 84.5\% & \textbf{85.8\%} \\
        + elastic transformation & 74.8\% & \textbf{79.0\%} & 85.0\% & 84.4\% \\
        + salt-and-pepper noise & 74.6\% & 78.2\% & 83.7\% & 85.5\% \\
        + Gaussian noise & \textbf{76.1\%} & 78.9\% & 84.9\% & 85.4\% \\
        + Poisson noise & 74.4\% & 76.4\% & 81.9\% & 83.5\% \\
        + solarize & 71.3\% & 76.1\% & 82.7\% & 85.1\% \\
        \hline
    \end{tabular*}
        \caption{\textbf{Accuracy of SwAV with ResNet50 backbone trained with an increasing number of augmentations from our augmentation pipeline on the test dataset.} Augmentations are added one after another, with each row adding one augmentation to those listed above. Augmentations are grouped by categories (geometric, brightness/contrast, sharpness/blur, deformation, noise, color). The best result for each evaluation metric is marked in bold. Note that the results using all augmentations correspond to the results of SwAV with a ResNet50 backbone from the second test step (cf. Figure~\ref{fig:selfsupervised} and Supplementary Table~\ref{tab:selfsupervised}).}
        \label{tab:augmentations}
\end{table}
 
\subsection*{Step 5: Transfer to a new setting}
\begin{figure}[t!]
\centering
\includegraphics[width=\linewidth]{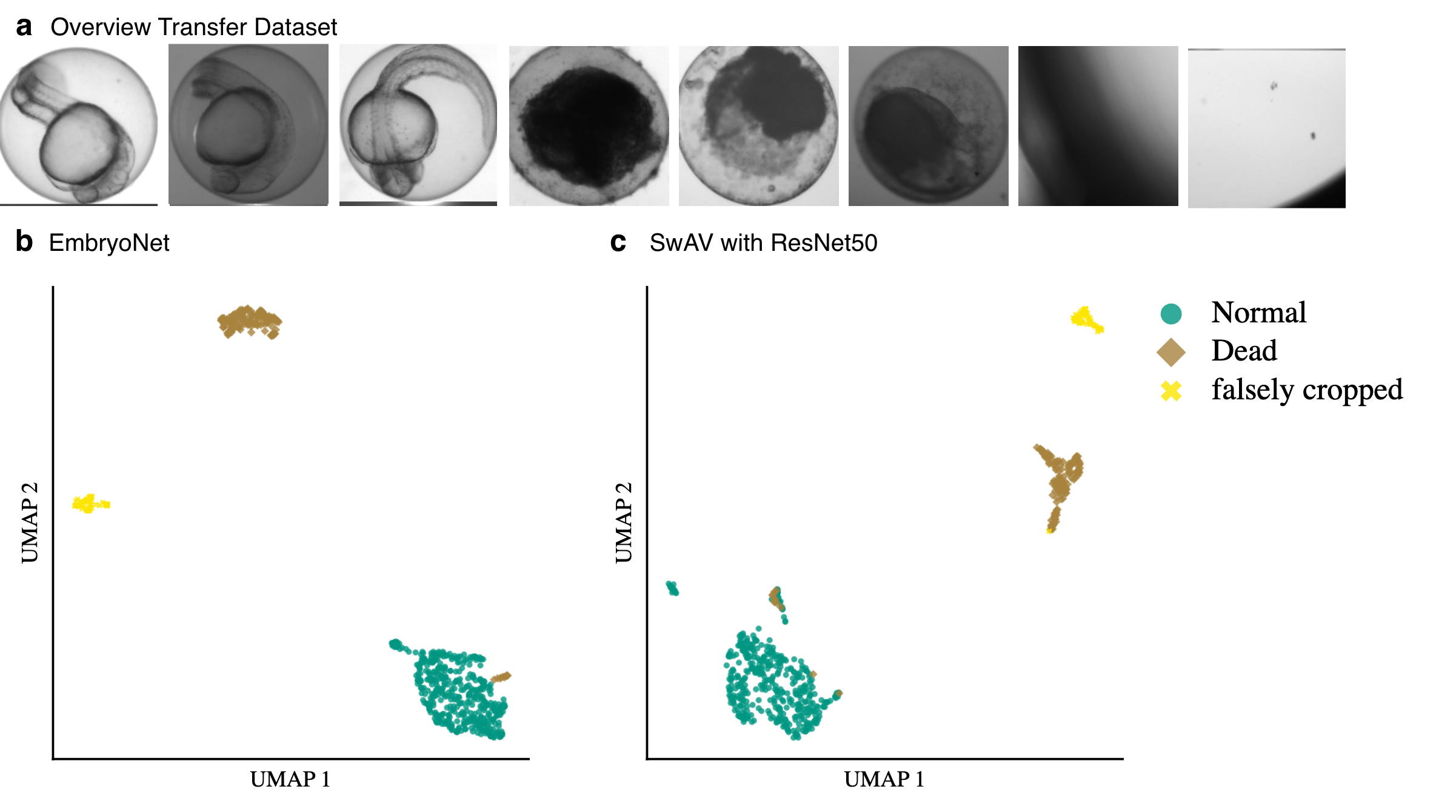}
\caption{\textbf{Example images and UMAPs~\cite{mcinnes2018umap} from the transfer dataset.} Example images from the transfer dataset are shown in (\textbf{a}): from left to right: three normal, three dead, two falsely cropped. UMAP~\cite{mcinnes2018umap} projections of the representations of the published EmbryoNet model~\cite{embryonetdata} with the classification head removed are shown in (\textbf{b}). UMAP projections of the learned representations of SwAV with a ResNet50 backbone trained for 800 epochs are depicted in (\textbf{c}).}
\label{fig:transfer_umaps}
\end{figure}

In the fifth test step, we test generalization to a new setting using 696 brightfield images of zebrafish embryos at 24 \gls{hpf}. Unlike the original dataset, the images were acquired with a Leica THUNDER Imager in a different laboratory (example images in Figure~\ref{fig:transfer_umaps}a). The dataset features normal and dead embryos. In accordance with the preprocessing of the EmbryoNet dataset~\cite{capek2023embryonet}, we extract embryo crops with the same classical circle-detection algorithm (Hough circle transform)~\cite{duda1972use}. For 58 images, the embryo was not detected correctly. We therefore form a third class of falsely cropped embryos, resulting in a three-class problem. This is to simulate a fully automated image-processing system. In practice, these embryos could be excluded, cropped manually, or a better detection algorithm could be applied. The resulting dataset features 472 normal, 166 dead, and 58 falsely cropped embryos. 

We use the overall best-performing model, SwAV with a ResNet50 backbone trained for 800 epochs on the EmbryoNet dataset, and compare it to the published EmbryoNet model~\cite{embryonetdata}. We evaluate two transfer strategies. In the first strategy, we do not retrain the models. Instead, we reuse the frozen linear classifier from the third step of the study plan. For the EmbryoNet model, we use the full model including the classification head. In the second strategy, we remove the classification head of the EmbryoNet model and train an XGBoost~\cite{chen2016xgboost} classifier on top of the frozen features of both the SwAV and the EmbryoNet model. While the whole dataset can be used directly for testing in the first scenario, we use five-fold cross-validation for evaluation in the second scenario.

Without retraining, the frozen linear classifier achieves 81.3\% accuracy across all images. The published EmbryoNet model reaches 75.9\%  accuracy. Since these models still predict the EmbryoNet phenotypes, we map the normal and dead classes to the respective EmbryoNet classes and the falsely cropped class to the unknown class. Every prediction of another phenotype counts as a false classification. With XGBoost trained on the learned representations of SwAV, we reach an accuracy of 98.3\%. Using the representations of the EmbryoNet model for XGBoost training, result in an accuracy of 99.1\%. The \gls{umap} visualizations of the learned representations are given in  Figure~\ref{fig:transfer_umaps}b \& c.
 
\section*{Discussion}

Adapting the evaluation to a single, fixed developmental stage matches how most zebrafish laboratories operate. This constraint lowers the accuracy of the published EmbryoNet model relative to the original report~\cite{capek2023embryonet}. By choosing a suitable \textbf{supervised architecture} and training on images from a similar developmental stage, the performance of EmbryoNet can be exceeded. CutMix further improves accuracy, yielding an overall improvement of nearly six percentage points. In general, smaller models ranging from 9M to 50M parameters perform best. Larger models did not automatically improve results, likely because of limited training data. All transformer-based architectures perform worse than \glspl{cnn} across the board. \Glspl{vit} typically require pretraining on tens to hundreds of millions of images to match \gls{cnn} performance, due to lack of locality and translation-equivariance biases inherent to \glspl{cnn}, which promote learning from limited data~\cite{dosovitskiy2021image}. The resulting weaker performance of transformers on a dataset of this size motivates restricting the \gls{ssl} comparison to \gls{cnn}-based methods. 

For our \textbf{comparison of \gls{ssl} algorithms}, SwAV performs best with a ResNet50 backbone, followed by MoCov2 and SimCLR. DINO performs poorly with the evaluated hyperparameters. The reason for this performance difference remains unclear. Linear probing on the frozen SwAV representations approaches the accuracy achieved by the published, fully supervised EmbryoNet model~\cite{embryonetdata} on the same test dataset, demonstrating the quality of the learned representations without using labels in the backbone training. 

From the different \textbf{pretraining strategies}, using only images from EmbryoNet emerges as the best approach. Longer pretraining improves performance. This behavior is well known from large natural-image datasets and holds for the domain-specific dataset in our investigation. Neither pretraining on \gls{imagenet} nor the naive combination of sequential pretraining on \gls{imagenet} and EmbryoNet achieves higher accuracies than the EmbryoNet-only approach given our pretraining configurations. More elaborate strategies for combining datasets could be explored in future work. For SwAV, the model organizes the representations mainly around the imaging device rather than the phenotype. However, the phenotypic information remains preserved as smaller variations within the per-microscope projections.

Finetuning the \gls{ssl}-pretrained model shows substantial improvements over supervised training with limited labeled data. Smaller standard deviations indicate higher robustness across different learning rates, which potentially leads to lower costs of hyperparameter searches, i.e., the trial-and-error tuning of training settings such as the learning rate. With all labels, finetuned SSL models and supervised training from scratch reach similar accuracy in the same protocol (means 88.0\%--88.9\%).

The \textbf{augmentation study} shows a clear saturation effect. The first few augmentations added improve results considerably, while augmentations beyond a certain point barely change the results. Overall, the number of augmentations appears to be more decisive than the application of specific augmentations. 

The \textbf{transfer experiment} illustrates the problem with reusing trained models across laboratories. The performance of both the SwAV model and the published EmbryoNet model~\cite{embryonetdata} drops meaningfully. However, without retraining SwAV provides a stronger starting point for a new laboratory and microscope. This is the case even though SwAV was trained only on images from 25 to 26 \gls{hpf}, while the test set was recorded at 24 \gls{hpf}. Using a lightweight classifier on top of the frozen features can achieve high accuracy with only a small training dataset. This holds for both models, with EmbryoNet~\cite {embryonetdata} slightly outperforming SwAV. Once enough labeled data is collected in the new setting, finetuning is also an option. 

Images acquired by fully automated screening devices will likewise differ from images taken under standard laboratory conditions. \Gls{ssl} models appear well suited for initial use in such settings because of better performance without retraining compared to the supervised model tested here. Through better transferability between laboratories, compliance with the 3R principles~\cite{russell1959principles} can be increased as fewer zebrafish embryos are needed for model adaptation.

Overall, the main value of \gls{ssl} models for zebrafish embryo images lies in label-efficiency and adaptability without retraining to new settings, rather than in surpassing supervised training when high amounts of labeled data are available. Several \textbf{limitations} remain: Comparisons of \gls{ssl} algorithms rely on single training runs. Small differences between the results should therefore be interpreted with caution. The dependence of the SwAV representations on the imaging device also warrants explicit attention when models are reused across laboratories, although adaptation with a lightweight classifier seems feasible. Further, only one timepoint was used for evaluation. While the phenotypes were most prominent for the 26 \gls{hpf} timepoint in the EmbryoNet paper (as shown in Extended Data Fig. 6 in ref. \cite{capek2023embryonet}), this does not necessarily mean that the results transfer directly to other timepoints. In contrast to Oktay et al.~\cite{oktay2026self}, we train only on one dataset. This allows us to use the more varied abnormal phenotypes of the EmbryoNet dataset, which are excluded from training by Oktay et al.~\cite{oktay2026self}. Further, more comprehensive label-efficiency evaluations are possible on only one dataset.

For practitioners, we recommend the following approach. A ResNet50 backbone provides a strong default. More recent backbones, such as ConvNeXt-v2, may increase accuracy but require extensive hyperparameter tuning. The contrastive methods SimCLR and MoCov2 remain strong candidates alongside SwAV. Pretraining should use the domain dataset for extended periods if the compute budget allows it. Training pipelines should include a sufficient set of augmentations up to a point of saturation. For adaptation to a new setting, training a lightweight classifier on the frozen features of a pretrained backbone is a useful first step before considering finetuning.

\section*{Methods}
 
\subsection*{Dataset}
We use the EmbryoNet dataset~\cite{embryonetdata}, which contains ten zebrafish embryo phenotypes and is designed for classification. Seven phenotypes result from compounds that modify a specific developmental pathway. The remaining three cover normal development, dead embryos, and an unknown category for embryos whose phenotype the annotator could not determine. Two microscopes, an ACQUIFER Imaging Machine and a Keyence BZ-X810, were used for image acquisition~\cite{capek2023embryonet}. The dataset spans development from 2 to 26 \gls{hpf}, with each well approximately imaged every two minutes. Because toxicity tests are usually evaluated at one or a few timepoints~\cite{OECD2013FET}, we restrict the evaluation to the endpoint at 26 \gls{hpf}, where the majority of phenotypes are most pronounced (Extended Data Fig. 6 in ref. \cite{capek2023embryonet}). We adopt the predefined split between training data and independent test data. From the predefined training set, we use images from 25 to 26 \gls{hpf} to obtain a sufficient number of images at a comparable developmental stage. We split the training set well-wise into training and validation data with a 0.9 to 0.1 ratio, so all images from a given well and all timesteps from a given embryo remain in the same split. This yields overall 151,163 images, of which 135,484 form the training set and 15,679 the validation set. The images are derived from 3,742 embryos. The test dataset used for all evaluations consists of a single 26 \gls{hpf} image of each embryo, in total 781 images from the same number of embryos. Supplementary Figure~\ref{fig:datasplit} illustrates the dataset handling. For the finetuning experiment in the third step, we add wells one at a time until the desired fraction of the training set is reached. This well-wise addition mimics a realistic scenario where an experiment uses fewer wells but labels all available images. A class-weighted loss, which gives rare phenotypes more weight during training, is used throughout all supervised trainings and finetunings to account for the dataset's class imbalance.
 
\subsection*{Study plan}
The study follows a five-step plan in which the performance of the models in earlier steps determines the models used in later steps. An overview of the study plan is given in Supplementary Figure~\ref{fig:testplan}. The study plan was mainly designed to avoid combinatorial explosion and extensive usage of computational resources. The first step compares network architectures trained in a fully supervised manner. The second step uses the best architectures as backbones to benchmark \gls{ssl} algorithms. The third step explores pretraining strategies and label-efficient classification, using the best backbone and algorithm. The fourth step examines the influence of augmentations in the augmentation pipeline, and the fifth step tests transfer to a new laboratory and microscope. Unless otherwise noted, the training setups follow the original papers of the respective methods~\cite{he2016deep, liu2022convnet, woo2023convnext, dosovitskiy2021image,liu2022swin, chen2020simple, chen2020improved, grill2020bootstrap, zbontar2021barlow, caron2020unsupervised, caron2021emerging}. All training runs are seeded.
 
\subsection*{Step 1: Supervised architecture comparison}
The supervised models are trained from scratch on the full training set. The best checkpoint is chosen based on the accuracy reached on the validation set. As an additional baseline, we include the EmbryoNet architecture without the time-channel, i.e., the additional input of the embryo age, which carries no information in our single-timepoint setup. The tested architectures come from the ResNet family~\cite{he2016deep}, the ConvNeXt-v1~\cite{liu2022convnet} and ConvNeXt-v2~\cite{woo2023convnext} families, vanilla \glspl{vit}~\cite{dosovitskiy2021image}, and Swin-v2 transformers~\cite{liu2022swin}. For the best architectures, we add CutMix~\cite{yun2019cutmix} to the augmentations, which improved performance in preliminary tests. For the published EmbryoNet model, the time-channel input is set to 1, which corresponds to 26 \gls{hpf}. We apply the same test-time augmentations for the EmbryoNet evaluation as in the original setup~\cite{capek2023embryonet}. Swin-v2 window sizes are set to 14, 14, 14, and 7 in one configuration (denoted as w14) and to 7 throughout in a second configuration (denoted as w7).
 
\subsection*{Step 2: Self-supervised algorithm comparison}
We compare state-of-the-art \gls{cnn}-based \gls{ssl} algorithms, namely SimCLR~\cite{chen2020simple}, MoCov2~\cite{chen2020improved}, BYOL~\cite{grill2020bootstrap}, Barlow Twins~\cite{zbontar2021barlow}, SwAV~\cite{caron2020unsupervised}, and DINO~\cite{caron2021emerging}, each trained for 200 epochs with ResNet50 and ConvNeXt-v2 pico backbones. A brief description of each algorithm's auxiliary task is provided in the Supplementary Methods. Approaches that are mainly designed for \glspl{vit}, such as MoCov3~\cite{chen2021empirical}, DINOv2~\cite{oquab2023dinov2}, masked autoencoders~\cite{he2022masked}, and I-JEPA~\cite{assran2023self}, are excluded due to the need of extensive pretraining data of \glspl{vit} and the poor performance of transformer-based architectures in the supervised testing during Step 1.
 
The quality of the learned representations is assessed on the frozen backbone using \gls{knn} classifications with k of 1, 5, 10, 20, and 200 neighbors, linear probing, and non-linear probing with an \gls{mlp} head. The representations are L2-normalized for the \gls{knn} evaluation, and the full training dataset is used. Linear probing follows the procedure described in the original SimCLR paper~\cite{chen2020simple}. The \gls{mlp} head consists of two fully connected layers with the same dimensionality as the learned features, and a final classification layer with 10 nodes, one per class. ReLU is applied between layers and softmax after the classification layer. Training is done using the same procedure as linear probing. Both probing variants are trained on the full training set with best-checkpoint selection on the validation set.
 
\subsection*{Step 3: Pretraining strategies and label-efficient classification}
Three pretraining strategies are compared for SwAV with a ResNet50 backbone: training on \gls{imagenet} for 800 epochs, training on EmbryoNet for 800 epochs, and training on \gls{imagenet} for 800 epochs followed by 200 epochs on EmbryoNet. A fully supervised ResNet50 trained on \gls{imagenet} from the MMPretrain model zoo~\cite{mmpretrain} serves as a baseline. For the sequential strategy, the stem and first residual stage, i.e., the low-level edge and texture filters, corresponding to the first 10 convolutional layers, of the ResNet50 are kept frozen. This includes the respective BatchNorm layers. For each epoch of the subsequent EmbryoNet training, we run \gls{knn} with k = 200 neighbors on the full validation dataset and use the best-performing backbone for the evaluations. As in the second step, we use \gls{knn}, linear probing, and non-linear probing via an \gls{mlp} head for evaluation of the representations. UMAP visualizations of the learned representations were computed with 30 nearest neighbors, a minimum distance of 0.1, and cosine distance in the input space.

For label-efficient classification, each pretrained model is finetuned with 5\%, 10\%, 20\%, 50\%, and 100\% of the labeled training data. The stem and first residual stage of the ResNet50 are kept frozen for all models during finetuning. A ResNet50 without frozen layers is trained from scratch on the same fractions as an additional baseline. The validation set is reduced by the same fraction as the training set. For each condition, five finetuning runs are performed that multiply the base learning rate of ResNet50~\cite{he2016deep} (0.1, with an SGD optimizer) by 0.1, 0.25, 0.5, 0.75, and 1.0, while the classification head keeps the base learning rate. Accordingly, the reported standard deviations describe the sensitivity to the learning rate rather than run-to-run noise.
 
\subsection*{Step 4: Augmentation study}
For the supervised training, the \gls{ssl} training, and the probing evaluations, we use the same augmentations as the original EmbryoNet paper~\cite{capek2023embryonet}, namely cropping, horizontal flip, rotation, random brightness and contrast changes, random gamma correction, sharpening, motion blur, salt-and-pepper noise, grid distortion, and optical distortion. Additionally, to include a wide range of augmentations in the augmentation pipeline, we also apply contrast-limited adaptive histogram equalization (CLAHE), defocus blur, elastic transformation, Gaussian noise, Poisson noise, and solarization. All augmentations with parameters are listed in Supplementary Table \ref{tab:augmentations_list}. The augmentation study for the fourth step starts with cropping and adds augmentations one after another, with each row of Table~\ref{tab:augmentations} adding one to those already applied.
 
\subsection*{Step 5: Transfer to a new setting}
For the transfer experiment, 696 brightfield images of individual zebrafish embryos at 24 \gls{hpf} were acquired with a Leica THUNDER Imager using a 2.5$\times$ objective, with the embryos kept in 96 round-bottom well plates in 250$\mu$l of E3 medium at 28$^{\circ}$C. In total, there are 128 control embryos without treatment, 120 embryos treated with varying concentrations of iota-carrageenan, 120 embryos treated with varying concentrations of lambda-carrageenan, 120 embryos treated with varying concentrations of poly-epsilon-lysine, 48 embryos treated with varying concentrations of polyhexamethylene biguanide, 80 embryos treated with 4.5\% ethanol, and 80 embryos treated with 0.2\% tricaine. Detailed descriptions of the individual treatments are given in Supplementary Table~\ref{tab:compounds}. All zebrafish husbandry and handling were performed in compliance with the German animal protection standards and approved by the Government of Baden-W\"{u}rttemberg, Regierungspr\"{a}sidium Karlsruhe, Germany (Aktenzeichen 35-9185.64/BH KIT).

Each acquisition is a z-stack of 11 planes spaced 100$\mu$m apart, and the sharpest plane is selected based on the variance of the Laplacian, a standard sharpness measure. The dataset contains 486 normal and 210 dead embryos. Embryo crops are extracted with a Hough circle transform~\cite{duda1972use} as in the original EmbryoNet paper~\cite{capek2023embryonet}, and the images without a valid detection are assigned to a third class rather than discarded, which results in a three-class problem. The cropped dataset features 472 crops of normal embryos, 166 crops of dead embryos, and 58 that are falsely cropped.

The best model from the study plan, SwAV with a ResNet50 backbone pretrained for 800 epochs on EmbryoNet, is evaluated in two ways: first, with the frozen linear classifier from the third step of the study plan without any retraining, and second, with an XGBoost classifier trained on the frozen representations with five-fold cross-validation using the same splits for both models. A classification counts as correct when a normal embryo is assigned to the normal class, a dead embryo to the dead class, and a falsely cropped embryo to the unknown class. The same two evaluations are run with the published EmbryoNet model, using either the full model or training an XGBoost on the features. The time-channel input of the EmbryoNet model was set to 0.917, corresponding to 24 \gls{hpf}. To ensure comparability, we use the test-time augmentation setup from the original EmbryoNet~\cite{capek2023embryonet} for both models. XGBoost classifiers use 100 trees with a maximum depth of 4 and a learning rate of 0.1. The UMAP visualizations use the same parameters as in Step 3.
 
\subsection*{Implementation}
The supervised training, the SimCLR, MoCov2, BYOL, Barlow Twins, and SwAV pretraining, and the linear and \gls{mlp} evaluations use MMPretrain~\cite{mmpretrain} 1.2.0. The DINO pretraining uses lightly~\cite{lightly} 1.5.22. The \gls{knn} evaluation follows the script of the original DINO repository~\cite{caron2021emerging}. The image augmentations use Albumentations~\cite{albumentations} 2.0.8. All trainings and evaluations were run on the HoreKa HPC cluster using either NVIDIA A100 (40 GB), H100 (94 GB), or H200 GPUs.

\bibliography{literature}

\section*{Acknowledgements}
This work has received funding from the European Union's HORIZON-CL4-2023-RESILIENCE-01 Research and Innovation programme under Grant Agreement No 101138387 (TOXBOX). Views and opinions expressed are however those of the authors only and do not necessarily reflect those of the European Union. The European Union cannot be held responsible for them.\\
The present contribution is supported by the Helmholtz Association under the joint research school "HIDSS4Health - Helmholtz Information and Data Science School for Health." and the program "Natural, Artificial and Cognitive Information Processing (NACIP)".\\
The authors gratefully acknowledge the computing time provided on the high-performance computer HoreKa by the National High-Performance Computing Center at KIT (NHR@KIT). This center is jointly supported by the Federal Ministry of Education and Research and the Ministry of Science, Research and the Arts of Baden-Württemberg, as part of the National High-Performance Computing (NHR) joint funding program (https://www.nhr-verein.de/en/our-partners). HoreKa is partly funded by the German Research Foundation (DFG).\\
This research was also supported by the Ministry of Science, Technological Development and Innovation of the Republic of Serbia, contract numbers 451-03-34/2026-03/200108 (Faculty of Mechanical and Civil Engineering in Kraljevo, University of Kragujevac) and 451-03-33/202603/200107 (Faculty of Engineering, University of Kragujevac). This research is in line with the strategic plan of the Republic of Serbia 2030, particularly in the section 2.2.2. Development of human resources, goal 3: Good health.\\

\section*{Author contributions statement}
\textbf{Thomas Lautenschlager:} Conceptualization, Methodology, Software, Investigation, Writing - Original Draft, Writing - Review \& Editing. \textbf{Ga\"{e}lle Hayot:} Resources, Data Curation, Writing - Review \& Editing. \textbf{Quentin Brassart:} Resources, Data Curation, Writing - Review \& Editing. \textbf{Angelo Jovin Yamachui Sitcheu:} Methodology, Writing - Review \& Editing. \textbf{Linus Schilpp:} Methodology, Writing - Review \& Editing. \textbf{Tijana Geroski:} Writing - Review \& Editing. \textbf{Tetyana Goldanova:} Writing - Review \& Editing. \textbf{Stephan Allgeier:} Writing - Review \& Editing. \textbf{Thibaut Soubri\'{e}}: Writing - Review \& Editing. \textbf{Milica Nikolic:} Writing - Review \& Editing. \textbf{Thomas Dickmeis:} Resources, Data Curation, Writing - Review \& Editing, Funding acquisition. \textbf{Ralf Mikut:} Conceptualization, Methodology, Writing - Review \& Editing, Supervision, Project administration, Funding acquisition. \textbf{Nils Friederich:} Conceptualization, Methodology, Writing - Review \& Editing, Supervision, Project administration.

\section*{Code availability}
Code will be made available upon publication.

\section*{Data availability}
The original EmbryoNet dataset~\cite{embryonetdata} is available under \url{https://doi.org/10.48606/15}. The dataset used for the transfer to a new setting will be made available upon publication. A selection of the best-performing models will be made available upon publication.

\section*{Additional information}

\textbf{Competing interests}\\
The authors declare no competing interests.

\newpage

\section*{Supplementary Methods}
\setcounter{figure}{0}
\renewcommand{\thefigure}{S\arabic{figure}}
\renewcommand{\figurename}{Supplementary Figure}
\setcounter{table}{0}
\renewcommand{\thetable}{S\arabic{table}}
\renewcommand{\tablename}{Supplementary Table}

\subsection*{Description of the self-supervised algorithms}
SimCLR~\cite{chen2020simple} is a contrastive method that creates two augmented views of each image. The model learns to bring the two views closer in representation space while pushing views of different images in the same batch apart. This makes large batch sizes beneficial. MoCov2~\cite{chen2020improved} relies on the same contrastive principle but draws negative examples from a queue. A slowly updated momentum encoder populates this queue, effectively removing the strict need for large batches. BYOL~\cite{grill2020bootstrap} operates entirely without negative examples. The algorithm instead trains an online network to predict the representation produced by a momentum-updated target network for a second view. A predictor head and a stop-gradient operation prevent representation collapse. Barlow Twins~\cite{zbontar2021barlow} also uses no negatives. This method drives the cross-correlation matrix between the representations of the two views toward the identity matrix, thereby reducing redundancy across feature dimensions. SwAV~\cite{caron2020unsupervised} assigns the features of each view to a set of learned prototypes. The algorithm leverages online clustering and multi-crop augmentation to enforce that the two views of an image predict each other's cluster assignment. DINO~\cite{caron2021emerging} trains a student network to match the output distribution of a momentum-updated teacher over a set of prototypes. Centering and sharpening operations prevent collapse. The method was primarily designed for \glspl{vit}. 

\subsection*{Implementation details}
The ConvNeXt-v2 models follow the training protocol of ConvNeXt-v1~\cite{liu2022convnet}, because the original ConvNeXt-v2 publication specifies only the pretraining and finetuning hyperparameters~\cite{woo2023convnext}. Batch sizes are reduced to 1024 for ConvNeXt training and to 2048 for ViT training, with the learning rate adjusted by square-root scaling~\cite{hoffer2017train}. We adjust the window sizes for Swin-v2 to accommodate 224×224 images.

\newpage
\section*{Supplementary Results}

\begin{table}[h]
    \begin{tabular*}{\linewidth}{@{\extracolsep{\fill}}lcc}
     Backbone & Accuracy & Parameters \\
     \hline
        published EmbryoNet model~\cite{embryonetdata} & 87.1\% & 14.0M \\
         \hline
        EmbryoNet (without time-channel) & 87.3\% & 14.0M \\
        ResNet18 & 89.5\% & 11.7M \\
        ResNet50 & \textbf{89.9\%} & 25.6M \\
        ResNet101 & 89.8\% & 44.6M \\
        ResNet152 & 89.2\% & 60.2M \\
        ConvNeXt-v1 tiny & 88.5\% & 28.6M \\
        ConvNeXt-v1 small & 89.5\% & 50.2M \\
        ConvNeXt-v1 base & 89.2\% & 88.6M \\
        ConvNeXt-v2 pico & \textbf{89.9\%} & 9.1M \\
        ConvNeXt-v2 nano & 88.3\% & 15.6M \\
        ConvNeXt-v2 tiny & 88.5\% & 28.6M \\
        ConvNeXt-v2 base & 88.9\% & 88.7M \\
        ViT-B/16 & 84.8\% & 86.6M \\
        Swin-v2 tiny w14 & 86.2\% & 27.6M \\
        Swin-v2 small w14 & 86.8\% & 49.0M \\
        Swin-v2 base w14 & 88.0\% & 86.9M \\
        Swin-v2 base w7 & 87.5\% & 86.9M \\
        \hline
        ResNet50 (CutMix) & 90.3\% & 25.6M \\
        ResNet101 (CutMix) & 91.4\% & 44.6M \\
        ConvNeXt-v1 small (CutMix) & \textbf{92.6\%} & 50.2M \\
        ConvNeXt-v2 pico (CutMix) & 91.7\% & 9.1M \\
        \hline
    \end{tabular*}
    \caption{\textbf{Accuracy of the backbones trained fully supervised from scratch on the test dataset.} All backbones are trained on the full training set, with best-checkpoint selection on the validation set. Models that additionally use CutMix are marked accordingly. The best results are marked in bold. The published EmbryoNet model serves as a baseline. Accuracy and model size are depicted as a scatter plot in Figure~\ref{fig:scatter}.}
    \label{tab:supervised}
\end{table}

\begin{table}[t]
    \begin{tabular*}{\linewidth}{@{\extracolsep{\fill}}llccccccc}
     SSL approach & backbone & k-NN 1 & k-NN 5 & k-NN 10 & k-NN 20 & k-NN 200 & \makecell[l]{linear\\probing} & \makecell[l]{non-linear\\probing\\(MLP head)} \\
        \hline
         & \makecell[l]{published EmbryoNet\\model~\cite{embryonetdata}} & 80.8\% & 82.1\% & 82.5\% & 82.1\% & 82.3\% & 76.5\% & 88.3\% \\
         \hline
        SimCLR & ResNet50 & 71.7\% & 72.1\% & 72.3\% & 72.1\% & \textbf{76.7\%} & 80.8\% & 82.8\% \\
         & ConvNeXt-v2 pico & 68.6\% & 71.3\% & 71.8\% & 73.0\% & 75.9\% & 79.6\% & 81.2\% \\
        MoCov2 & ResNet50 & \textbf{73.8\%} & 72.9\% & 73.0\% & 73.8\% & 74.8\% & 81.9\% & 83.8\% \\
         & ConvNeXt-v2 pico & 70.3\% & 70.8\% & 71.2\% & 71.1\% & 73.9\% & 75.0\% & 78.1\% \\
        BYOL & ResNet50 & 69.6\% & 70.4\% & 70.3\% & 70.0\% & 71.3\% & 77.7\% & 81.8\% \\
         & ConvNeXt-v2 pico & 71.8\% & \textbf{73.5\%} & \textbf{73.2\%} & \textbf{74.0\%} & 75.3\% & 80.0\% & 80.7\% \\
        Barlow Twins & ResNet50 & 64.1\% & 64.0\% & 64.3\% & 64.0\% & 66.2\% & 74.8\% & 76.4\% \\
         & ConvNeXt-v2 pico & 58.1\% & 61.5\% & 61.5\% & 61.6\% & 63.5\% & 59.8\% & 69.4\% \\
        SwAV & ResNet50 & 68.8\% & 70.4\% & 70.4\% & 71.3\% & 76.1\% & \textbf{82.7\%} & \textbf{85.1\%} \\
         & ConvNeXt-v2 pico & 57.0\% & 57.5\% & 57.7\% & 57.4\% & 61.8\% & 63.9\% & 69.5\% \\
        DINO & ResNet50 & 44.2\% & 44.9\% & 44.8\% & 44.7\% & 51.1\% & 43.1\% & 51.6\% \\
         & ConvNeXt-v2 pico & 35.5\% & 38.2\% & 38.9\% & 39.8\% & 40.1\% & 34.4\% & 41.1\% \\
         \hline
    \end{tabular*}
    \caption{\textbf{Evaluation of the learned representations of the self-supervised algorithms with the respective backbones on the test dataset.} Accuracy is determined through k-NN classification for k of 1, 5, 10, 20, and 200, linear probing, and non-linear probing with an \gls{mlp} head on the frozen backbones. The best results among the SSL approaches are marked in bold. Results are visualized in Figure~\ref{fig:selfsupervised}.}
    \label{tab:selfsupervised}
\end{table}
 
\begin{table}[t]
    \begin{tabular*}{\linewidth}{@{\extracolsep{\fill}}lllcccccc}
     &\makecell[l]{PT\\IN1k} & \makecell[l]{PT\\EN} & 5\% labels & \makecell[l]{10\% labels} & \makecell[l]{20\% labels} & \makecell[l]{50\% labels} & \makecell[l]{100\% labels} &\\
     \hline
     \makecell[l]{supervised ResNet50} & no & no & 45.7\% (6.6) & 61.5\% (8.6) & 72.2\% (4.9) & 84.6\% (2.6) & 88.0\% (0.9)\\
     & yes (SV)& no & 27.1\% (5.2) & 57.0\% (15.9) & 75.4\% (4.5) & 84.5\% (3.1) & 88.6\% (1.0)\\
     SwAV ResNet50 & yes & no & 58.8\% (3.0) & 71.5\% (3.6) & 77.7\% (1.9) & 82.2\% (1.9) & 85.2\% (2.3)\\
     & no & yes & \textbf{70.2\%} (1.2) & \textbf{77.4\%} (0.6) & \textbf{81.7\%} (0.4) & \textbf{88.0\%} (0.3) & 88.7\% (0.4)\\
     & yes & yes & 66.8\% (1.3) & 77.1\% (1.8) & 80.4\% (0.4) & 84.9\% (0.5) & \textbf{88.9\%} (0.9)\\
     \hline
    \end{tabular*}
    \caption{\textbf{Accuracies (and standard deviations) for finetuning with different fractions of labeled data on the test dataset.} The results are reported as mean and standard deviation over five finetuning runs that are listed in Supplementary Table \ref{tab:all_runs} and differ only in the learning rate. SwAV with a ResNet50 backbone is pretrained with ImageNet-1k, EmbryoNet, or both, and a ResNet50 pretrained fully supervised on ImageNet-1k as well as a ResNet50 trained from scratch serve as baselines. The best results are marked in bold. Two of the five training runs with the supervised IN1k pretraining using 5\% labels did not converge. These runs were excluded from mean and standard deviation calculation. The standard deviations describe the sensitivity to the learning rate. PT denotes pretraining, EN denotes EmbryoNet, and SV denotes supervised. These results are visualized in Figure~\ref{fig:finetuning}.}
    \label{tab:finetuning}
\end{table}
 
\begin{table}[t]
    \begin{tabular*}{\linewidth}{@{\extracolsep{\fill}}lllcccccc}
     &\makecell[l]{PT\\IN1k} & \makecell[l]{PT\\EN} & \makecell[l]{5\%\\labels} & \makecell[l]{10\%\\labels} & \makecell[l]{20\%\\labels} & \makecell[l]{50\%\\labels} & \makecell[l]{100\%\\labels} &\\
     \hline
     supervised ResNet50 \\
     \\
     0.1$\times$ learning rate & no & no &35.2\% & 46.5\% & 64.0\% & 80.4\% & 86.6\%\\
     0.25$\times$ learning rate & no & no & 43.3\% & 62.0\% & 71.3\% & 84.3\% & 88.3\%\\
     0.5$\times$ learning rate & no & no & 48.8\% & 66.5\% & 75.4\% & 85.9\% & 88.0\%\\
     0.75$\times$ learning rate & no & no & 50.6\% & 65.6\% & 74.9\% & 85.1\% & 88.0\%\\
     1$\times$ learning rate & no & no & 50.4\% & 66.7\% & 75.5\% & 87.5\% & 89.2\%\\
     \\
     0.1$\times$ learning rate & yes (SV) & no & 30.6\% & 31.9\% & 76.3\% & 86.6\% & 88.2\% \\
     0.25$\times$ learning rate & yes (SV) & no & 29.6\% & 50.4\% & 78.6\% & 85.9\% & 88.5\% \\
     0.5$\times$ learning rate & yes (SV) & no & 21.1\% & 67.0\% & 71.4\% & 85.5\% & 89.5\% \\
     0.75$\times$ learning rate & yes (SV) & no & 0.0\%$^{\dagger}$ & 66.1\% & 70.0\% & 79.1\% & 89.5\% \\
     1$\times$ learning rate & yes (SV) & no & 0.0\%$^{\dagger}$ & 69.4\% & 80.5\% & 85.5\% & 87.3\% \\
     \\
     SwAV ResNet50 \\
     \\
     0.1$\times$ learning rate & yes & no & 57.9\% & 65.8\% & 74.6\% & 79.8\% & 81.7\%\\
     0.25$\times$ learning rate & yes & no & 58.8\% & 70.4\% & 77.5\% & 80.5\% & 84.5\%\\
     0.5$\times$ learning rate & yes & no & 58.6\% & 73.5\% & 78.2\% & 83.6\% & 85.1\%\\
     0.75$\times$ learning rate & yes & no & 63.5\% & 73.1\% & 78.2\% & 83.0\% & 87.8\%\\
     1$\times$ learning rate & yes & no & 55.3\% & 74.8\% & 79.9\% & 84.1\% & 86.7\%\\
     \\
     0.1$\times$ learning rate & no & yes & 68.5\% & 76.8\% & 81.8\% & 87.8\% & 88.7\% \\
     0.25$\times$ learning rate & no & yes & 69.7\% & 76.7\% & 81.6\% & 88.0\% & 88.2\% \\
     0.5$\times$ learning rate & no & yes & 70.4\% & 77.6\% & 82.3\% & 87.6\% & 89.4\% \\
     0.75$\times$ learning rate & no & yes & 71.2\% & 78.0\% & 81.2\% & 88.3\% & 88.7\% \\
     1$\times$ learning rate & no & yes & 71.3\% & 78.0\% & 81.8\% & 88.3\% & 88.6\% \\
     \\
     0.1$\times$ learning rate & yes & yes & 64.8\% & 74.4\% & 80.3\% & 84.5\% & 87.3\% \\
     0.25$\times$ learning rate & yes & yes & 66.3\% & 78.2\% & 80.5\% & 85.1\% & 88.9\% \\
     0.5$\times$ learning rate & yes & yes & 67.6\% & 76.1\% & 81.0\% & 85.3\% & 89.6\% \\
     0.75$\times$ learning rate & yes & yes & 68.2\% & 77.7\% & 79.9\% & 84.1\% & 89.2\% \\
     1$\times$ learning rate & yes & yes & 67.1\% & 79.0\% & 80.2\% & 85.3\% & 89.4\% \\
     \hline
    \end{tabular*}
    \caption{\textbf{Accuracies of the individual finetuning runs (5 runs each with different learning rates) for finetuning with different fractions of labeled data on the test dataset.} The results of the individual training runs underlie Figure~\ref{fig:finetuning} and Supplementary Table~\ref{tab:finetuning}. SwAV with a ResNet50 backbone is pretrained with ImageNet-1k, EmbryoNet, or both, and a ResNet50 pretrained fully supervised on ImageNet-1k as well as a ResNet50 trained from scratch serve as baselines. The used model is chosen with best-checkpoint selection on the validation set. PT denotes pretraining, EN denotes EmbryoNet, and SV denotes supervised. $^{\dagger}$ denotes runs that did not converge. Affected runs are excluded from mean and standard deviation calculation.}
    \label{tab:all_runs}
\end{table}

\begin{table}[t]
    \begin{tabular*}{\linewidth}{@{\extracolsep{\fill}}llcc}
    Augmentation Category & Transform & Parameters & Probability $p$ \\
    \hline
        \emph{Geometric} & & & \\
         & Random crop from borders & max crop width = 0.1 & 0.5 \\
         & Resize & height = 224, width = 224 & 1.0 \\
         & Horizontal flip & - & 0.5 \\
         & Rotation & range $\in$ [0, 360] & 1.0 \\
        \multicolumn{4}{l}{\emph{Brightness / Contrast} \quad (OneOf, $p=0.5$)} \\
         & \makecell[l]{Random brightness and\\contrast changes} & brightness $\in$ [-0.3, 0.3], contrast $\in$ [-0.3, 0.3] & 1.0 \\
         &  Random gamma correction & gamma $\in$ [83, 125] & 1.0 \\
         & CLAHE & clip range $\in$ [1, 4], tile grid = [8, 8] & 1.0 \\ 
        \multicolumn{4}{l}{\emph{Sharpness / Blur} \quad (OneOf, $p=0.5$)} \\
         & Sharpen & alpha $\in$ [0.1, 1.0], lightness $\in$ [0.5, 1.0] & 1.0 \\ 
         & Motion blur & blur limit $\in$ [3, 31] & 1.0 \\
         & Defocus blur & radius $\in$ [3, 10], alias blur $\in$ [0.1, 0.5] & 1.0 \\
        \multicolumn{4}{l}{\emph{Deformation} \quad (OneOf, $p=0.5$)} \\
         & Grid distortion & distortion $\in$ [-0.3, 0.3], steps = 5 & 1.0 \\
         & Optical distortion & distortion $\in$ [0, 0.75] & 1.0 \\
         & Elastic transformation & displacement $\in$ [0.02, 0.05], control grid = [7, 7] & 1.0 \\
        \multicolumn{4}{l}{\emph{Noise} \quad (OneOf, $p=0.5$)} \\
         & Salt-and-pepper noise & amount $\in$ [0.0, 0.02], salt vs. pepper = 0.5 & 1.0 \\
         & Gaussian noise & std $\in$ [0.0, 0.44], mean = 0 & 1.0 \\
         & Poisson noise & range $\in$ [0.1, 0.3] & 1.0 \\
        \multicolumn{4}{l}{\emph{Color}} \\
         & To gray transformation & - & 1.0 \\
         & Solarize & threshold = 0.5 & 0.5 \\
    \end{tabular*}
    \caption{\textbf{List of augmentations used in the augmentations pipeline with their parameters and probabilities.} For the categories denoted with "OneOf", only one augmentation is applied at a time with the given probability of "OneOf".\\
    Note: Vertical flip is not applied even though it is used in EmbryoNet~\cite{capek2023embryonet}, because of the redundancy between vertical and horizontal flips combined with 360-degree rotation.\\
    Note: To gray transformation is applied to restore grayscale image after Gaussian or Poisson noise is added.}
    \label{tab:augmentations_list}
\end{table}

\begin{table}[t]
    \begin{tabular*}{\linewidth}{@{\extracolsep{\fill}}llrrrr}
     Batch & Compound & Concentration & Alive embryos & Dead embryos & \makecell[l]{Total number\\of images} \\
     \hline
     1 & control & - & 23 & 1 & 24 \\
      & iota-carrageenan & 0.5 g/L & 24 & 0 & 24 \\
      & & 1.0 g/L & 24 & 0 & 24 \\
      & & 2.0 g/L & 24 & 0 & 24 \\
      & & 4.0 g/L & 24 & 0 & 24 \\
      & & 8.0 g/L & 24 & 0 & 24 \\
    2 & control & - & 24 & 0 & 24 \\
      & lambda-carrageenan & 0.5 g/L & 23 & 1 & 24 \\
      & & 1.0 g/L & 24 & 0 & 24 \\
      & & 2.0 g/L & 24 & 0 & 24 \\
      & & 4.0 g/L & 24 & 0 & 24 \\
      & & 8.0 g/L & 24 & 0 & 24 \\
    3 & control & - &24 & 0 & 24 \\
      & poly-epsilon-lysine & 0.000313 g/L & 24 & 0 & 24 \\
      & & 0.000625 g/L & 24 & 0 & 24 \\
      & & 0.00125 g/L & 24 & 0 & 24 \\
      & & 0.0025 g/L & 24 & 0 & 24 \\
      & & 0.005 g/L & 24 & 0 & 24 \\
    4 & control & & 24 & 0 & 24 \\
      & polyhexamethylene biguanide & 0.000313 g/L & 0 & 24 & 24 \\
      & & 0.000625 g/L & 0 & 24 & 24 \\
    5 & control & - & 32 & 0 & 32 \\
      & ethanol & 4.5\% & 0 & 80 & 80\\
      & tricaine & 0.2\% & 0 & 80 & 80 \\
     \hline
    \end{tabular*}
    \caption{\textbf{Description of the transfer dataset.} Five batches, each with its own control, were either treated with iota-carrageenan, lambda-carrageenan, poly-epsilon-lysine, polyhexamethylene biguanide, or, in the case of the fifth batch, with ethanol or tricaine.}
    \label{tab:compounds}
\end{table}
 
\begin{figure}[t]
\centering
\includegraphics[width=\linewidth]{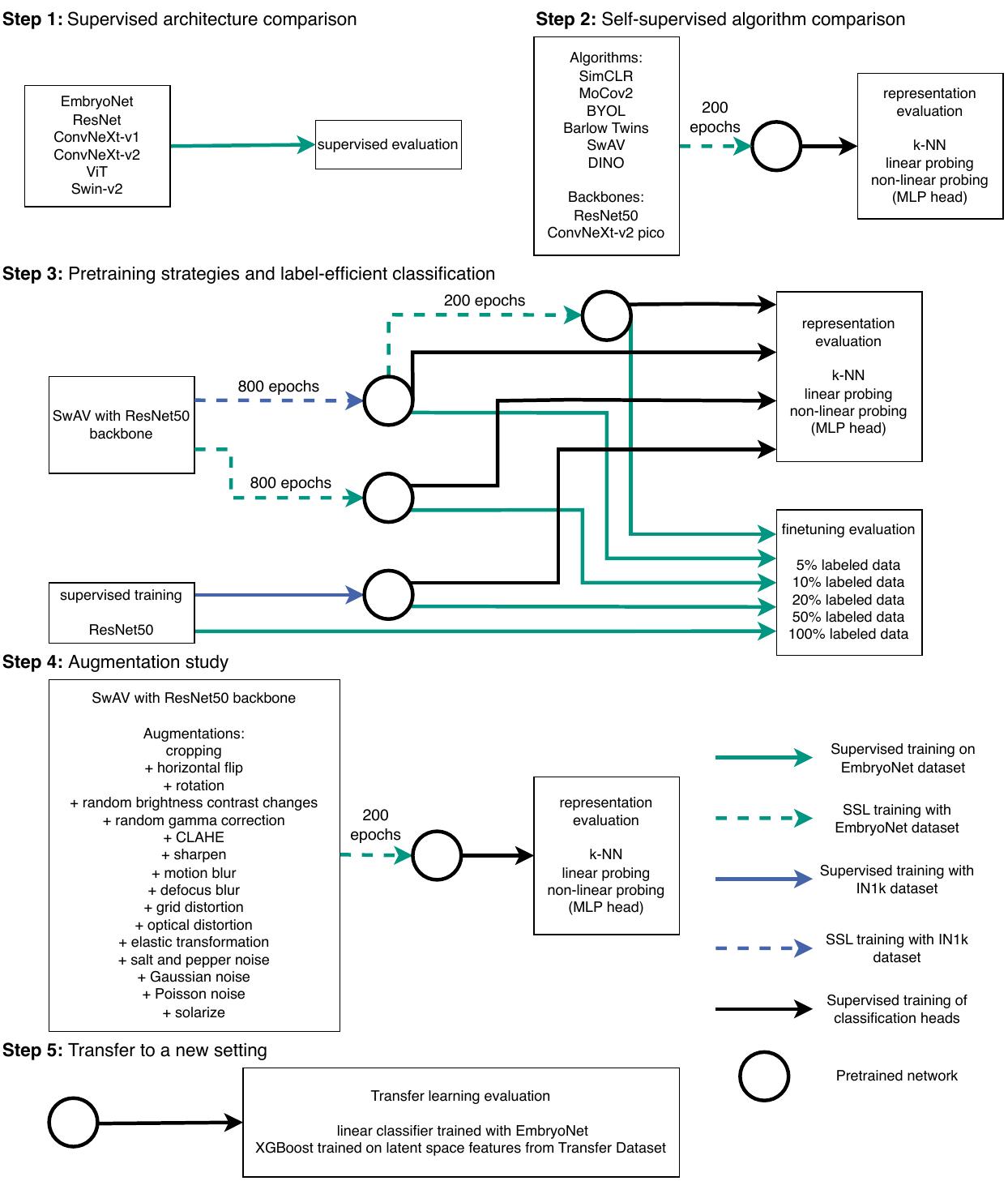}
\caption{\textbf{Visualization of the study plan.} \textbf{Step 1:} Different popular network architectures are tested fully supervised. \textbf{Step 2:} Various state-of-the-art CNN-based SSL algorithms are tested with ResNet50 and ConvNeXt-v2 pico as backbones. \textbf{Step 3:} Pretraining strategies using ImageNet-1k, EmbryoNet, or both sequentially are tested on SwAV with a ResNet50 backbone. \textbf{Step 4:} Augmentations from the augmentation pipeline are tested using SwAV with a ResNet50 backbone by adding one augmentation per training run to the already included ones. \textbf{Step 5:} A transfer study investigates how well a network pretrained with SSL can be used in a previously unseen setting.}
\label{fig:testplan}
\end{figure}
 
\begin{figure}[t]
\centering
\includegraphics[width=\linewidth]{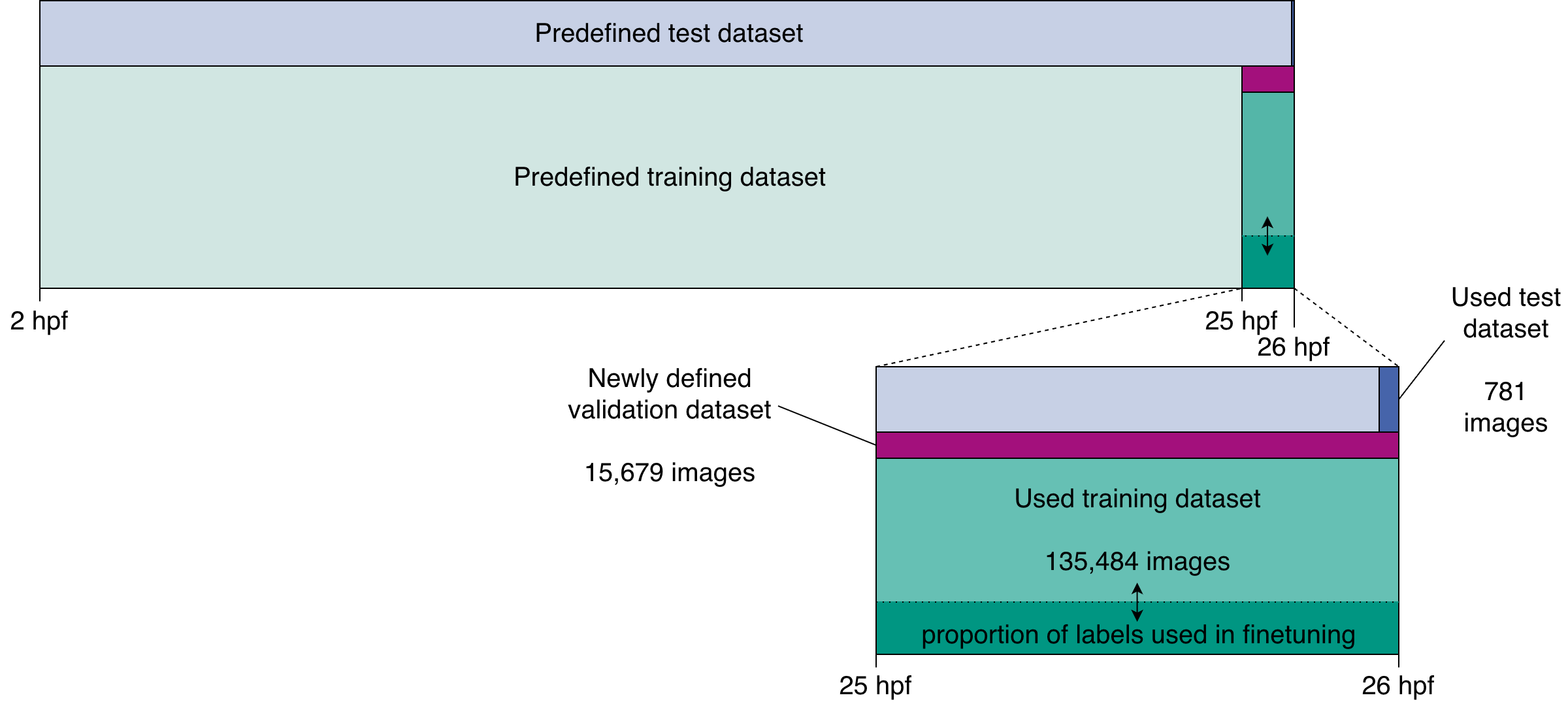}
\caption{\textbf{Handling of EmbryoNet dataset.} The predefined training (green) and test (blue) datasets from EmbryoNet are respected. Only the last images at 26 hpf of each embryo in the test dataset are used in all evaluations. Only images from 25 hpf to 26 hpf are used from the training dataset. A new validation dataset (purple) is defined, using 10\% of the original training dataset from 25 hpf to 26 hpf. The split is applied well-wise so that images of one embryo can either be in the training or the validation dataset. While SSL uses no labels for training, supervised training and training for the evaluation of the learned representations use all labels from the training dataset. The number of labels used for finetuning of SSL-pretrained backbones in the third step of the study plan varies between 5\% and 100\%. Proportions in the figure correspond to proportions of the actual datasets.}
\label{fig:datasplit}
\end{figure}

\end{document}